\documentclass{article}

\usepackage{iclr2027_conference,times}
\usepackage{amsmath,amssymb,bm}
\usepackage{booktabs,longtable,graphicx}
\usepackage{float,placeins}
\usepackage[table]{xcolor}
\definecolor{BoundedRow}{HTML}{EEF4FA}
\definecolor{BoundedAccent}{HTML}{25486B}
\definecolor{SceneTint}{HTML}{E8F0F8}
\definecolor{SceneRule}{HTML}{D0DAE4}
\newcommand{\SceneSeparator}{\arrayrulecolor{SceneRule}\specialrule{0.25pt}{0.65pt}{0.65pt}\arrayrulecolor{black}}
\usepackage{hyperref}
\usepackage{url}

\newcommand{\BoundedGSTitle}{BoundedGS: Memory-Efficient High-Resolution Training for 3D Gaussian Splatting}
\title{\raggedright\hyphenpenalty=10000\BoundedGSTitle}
\author{Xiaobin Deng \quad Changyu Diao\thanks{Corresponding authors.} \quad Min Li \quad Duanqing Xu\footnotemark[1]\\{\normalfont Zhejiang University}}
\iclrfinalcopy
\hypersetup{hidelinks,pdftitle={\BoundedGSTitle},pdfauthor={Xiaobin Deng, Changyu Diao, Min Li, Duanqing Xu}}

\begin{document}
\maketitle
\fancyhead{}
\renewcommand{\headrulewidth}{0pt}
\begin{abstract}
High-resolution 3D Gaussian Splatting training increases pixel and Gaussian--tile intersection counts, expanding rasterization caches that coexist with projection and gradient state at peak memory.
A complete-view update aggregates gradients across the view; each region's forward and backward caches can be released after its contribution is collected.
BoundedGS separates rasterization and projection backward passes and bounds the active local rasterization working set according to its measured load.
Candidate-indexed projection and compact gradient storage address the view-level peaks exposed by this change, while CPU-assisted structure rebuilding and bounded image supply manage resources outside rendering.
In a resolution sweep at a fixed count of about 6M Gaussians, the measured rendering working set stays within 1.073--1.077 GiB from 1K to 16K; at 11K, a warm pass takes 19.79 seconds versus 21.76 for full-frame execution.
Across five native 5K outdoor scenes, full training reduces the mean rendering working set from 9.97 to 1.76 GiB (82.3\%), retains similar reconstruction quality, and shortens training time from 36.7 to 33.9 minutes.
\par\noindent Project page: {\urlstyle{same}\url{https://xiaobin2001.github.io/BoundedGS-Web/}}.
\end{abstract}

\section{Introduction}

Training 3D Gaussian Splatting (3DGS)~\citep{kerbl20233d} at larger scale involves two distinct growth factors: scene extent and observation resolution.
Larger scenes commonly contain more Gaussians spread over a wider area, and spatial partitioning can distribute their training cost~\citep{lin2024vast,liu2025city}.
Higher observation resolution increases the resources required by each training view even when the scene extent and Gaussian count remain approximately fixed.
Extra computation can often be accommodated by a longer run, whereas a memory peak beyond device capacity prevents that view from being trained.
InfiniGS~\citep{liu2026infinigs}, for example, uses image blocks and asynchronous optimization to relieve the memory pressure of high-resolution training.
For a complete-view update on one GPU, the peak also depends on the simultaneous residency of rasterization, projection, and gradient state.
We study how to control this active working set as resolution grows while retaining efficient execution.

Decomposing the training step reveals that its rendering intermediates need not remain resident across the entire view.
A complete-view update requires the contributions from all image regions to be aggregated before the parameter update; it does not require every region's forward and backward computations to advance in lockstep.
A conventional path renders the entire view before computing the loss and backward pass, keeping the intermediate states needed for regional backward computation alive together.
Rasterization, however, is local to the image: given projected Gaussian attributes and local image gradients, a tile's forward and backward passes use its own ordered Gaussian list and pixel state.
While the Gaussian parameters remain fixed, the regional gradients can be computed separately and accumulated before the update.
Each region can therefore complete its loss and backward pass and release its local state before the next region runs.
Increasing resolution still creates more pixels and intersections, but limiting the simultaneously active local state controls the growth of the main rendering caches.

We introduce \textbf{BoundedGS}, a bounded-working-set execution method for high-resolution 3DGS training.
Within rasterization, it completes forward and backward computation region by region, releases intersections, compositing and loss state after extracting each gradient contribution, and chooses region sizes from their measured load.
For attribute evaluation and projection, it conservatively selects candidate Gaussians and can process projection groups in sequence, limiting the amount of detailed state generated from the global model.
The same candidates index compact parameter-gradient storage instead of allocating complete gradient rows for Gaussians unrelated to the current view.
All regions use one global Gaussian model and the original camera, cover every pixel, combine gradients and density-control statistics by view, and then perform one parameter update.
The memory savings arise from the range and lifetime of computations.
Once rendering peaks are reduced, structure updates and image supply can become the limiting resources; BoundedGS manages their temporary state as part of the full training system.

In a resolution sweep with the Gaussian count fixed at about 6M, the rendering working set remains within 1.073--1.077 GiB from 1K to 16K, while the warm pass at the shared 11K setting is about 9\% faster than full-frame execution.
Full training on five native 5K outdoor scenes reduces the rendering working set by 82.3\%, with similar final quality and slightly shorter training time.
Our contributions are:
\begin{itemize}
\item We identify the memory peak caused by co-resident rasterization, projection, and gradient state in complete-view training, and separate rasterization from projection backward computation so that local state can be released after its gradient contribution is collected.
\item We develop a load-bounded raster-leaf execution path and integrate candidate-indexed projection, compact gradients, structure rebuilding, and image supply into a single-GPU training system.
\item We use a 1K--16K resolution sweep at a fixed Gaussian count, component diagnostics, and full training on native high-resolution scenes to verify a nearly constant rendering working set, similar reconstruction quality, and training-time performance.
\end{itemize}

\section{Related Work}

\paragraph{Representation quality, model capacity, and training efficiency.}
3DGS~\citep{kerbl20233d} combines explicit Gaussians, adaptive densification, and differentiable rasterization for high-quality real-time rendering.
Subsequent work improves initialization, densification, reconstruction quality, and convergence~\citep{kotovenko2026edgs,ye2024absgs,kheradmand2024mcmc,deng2025improved}; reduces model capacity or storage through simplification, pruning, and encoding~\citep{fang2024mini,morgenstern2025sogs,chen2024hac,deng2025gns}; extends the primitive or representation to structured and dynamic settings~\citep{huang2024twodgs,wu2024fourdgs,lu2024scaffold}; and scales scene extent with spatial partitions and hierarchies~\citep{lin2024vast,liu2025city,kerbl2024hierarchical}.
We focus on the temporary memory peak of a training view while retaining the global model and input resolution.

\paragraph{Rasterization, projection, and intermediate state.}
Changing the rasterizer's computation and data organization is an established route to greater efficiency.
gsplat~\citep{ye2025gsplat} reduces training memory with packed projection state and optional sparse gradients for positions, rotations, and scales in its optimized CUDA rasterizer.
StopThePop~\citep{radl2024stopthepop} improves view consistency by hierarchical reordering and culling, changing visibility and compositing order.
FlashGS~\citep{feng2025flash}, Speedy-Splat~\citep{hanson2025speedy}, and AdR-Gaussian~\citep{wang2024adr} reduce Gaussian--tile work through precise overlap tests, opacity-aware ranges, and early culling.
Taming-3DGS~\citep{mallick2024taming} uses compositing checkpoints for Gaussian-parallel backward computation, while Faster-GS~\citep{hahlbohm2026faster} combines bounding, sorting, kernel fusion, backward, and optimizer improvements.
LiteGS~\citep{liao2025litegs} combines spatial clustering, selection, compact layouts, warp-level gradient reduction, and modified densification.
GS-TG~\citep{jo2025gstg} shares sorting across groups of tiles in a rendering accelerator and retains fine-grained participation of smaller tiles through bit masks.
These works improve training or rendering efficiency through compact states, overlap culling, sorting, and backward kernels.
BoundedGS targets the temporary training-memory peak by controlling which intermediate states coexist, while managing the time cost of local execution.

\paragraph{Scaling resources for high-resolution training.}
Prior work addresses high-resolution training through representation size, supervision sampling, and resource distribution.
EfficientGS~\citep{liu2025efficientgs} reduces redundancy through selective densification, pruning, and sparse growth of spherical-harmonic order.
Turbo-GS~\citep{dhiman2026turbo} renders a subset of pixels selected by dilated sampling for each supervision step and uses a convergence-aware growth budget.
HRGS~\citep{li2025hrgs} starts from a low-resolution global prior and partitions models and observations for high-resolution local refinement and fusion.
RetinaGS~\citep{li2024retinags} distributes Gaussian storage through model parallelism and combines the rendered contributions of submodels.
GGRt~\citep{li2024ggrt} performs a full-image forward pass without automatic differentiation and then uses cached image gradients to recompute backward contributions in blocks for generalizable reconstruction.
Grendel~\citep{zhao2025grendel} distributes Gaussian and pixel work across GPUs, dynamically balances load, combines gradients through communication, and supports multi-view batches.
InfiniGS~\citep{liu2026infinigs} combines multi-resolution training with image-block partitioning and asynchronous optimization, adjusting Adam hyperparameters and densification thresholds for resolution and partition changes.
These systems organize high-resolution views through multi-GPU pixel partitions or image blocks; we examine the lifetime of rasterization forward and backward caches within a complete view on one GPU.
CLM~\citep{zhao2026clm} and GS-Scale~\citep{lee2025gsscale} extend model capacity through host offloading, and CLM also reduces projection input and related gradient state by selecting frustum candidates.
BoundedGS combines view-level candidates with load-bounded local rasterization to control the simultaneous residency of Gaussian--tile intersections, pixels, projection state, and parameter gradients.

Together, these works address high-resolution training through supervision resolution, pixel sampling, model budgets, image blocks, and candidate filtering.
BoundedGS retains full-pixel supervision and one update per view, making resolution-dependent rasterization state an active bounded working set and integrating projection, gradients, and other resource management into the same training path.

\section{Method}

\subsection{The training step and memory lifetimes}

3DGS optimizes Gaussian parameters from training images.
For each camera, projection activates opacity, scale, and rotation, computes screen position, depth, and shape, and evaluates spherical-harmonic color.
We include these computations and their backward pass in the \emph{projection stage}.
The standard path typically processes all Gaussians, although each projection depends only on one Gaussian and the camera.

Rasterization expands Gaussian intersections with $16\times16$-pixel tiles, sorts them by tile and depth, and alpha-composites contributions.
For pixel $p$ with ordered contributors $\mathcal I_p$, its color is
\begin{equation}
\mathbf C_p
=\sum_{i\in\mathcal I_p}
\mathbf c_i a_{i,p}
\prod_{\substack{j\in\mathcal I_p\\j\prec i}}(1-a_{j,p})
+\mathbf c_{\mathrm{bg}}
\prod_{i\in\mathcal I_p}(1-a_{i,p}).
\end{equation}
Here $\mathbf c_i$ is view-dependent color, $a_{i,p}$ effective opacity, $j\prec i$ compositing order, and $\mathbf c_{\mathrm{bg}}$ background color.
Each pixel uses its own ordered contributors.

\begin{figure}[!htbp]
\centering
\includegraphics[width=\linewidth]{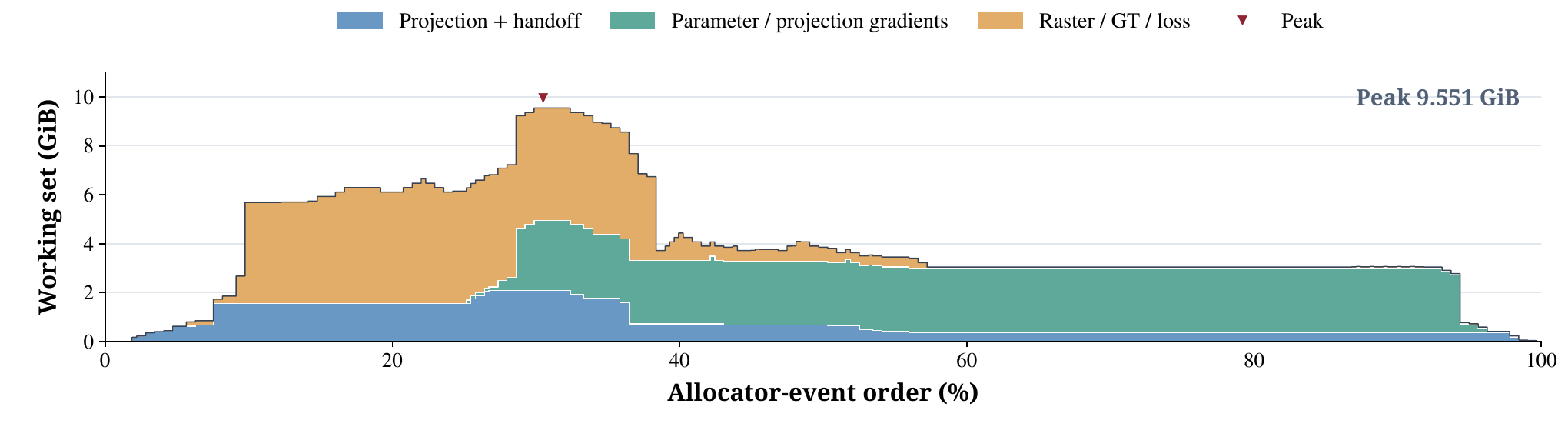}
\caption{Original working-set trace for one view of Bicycle's 15K SH3 model. The event-order axis and stacked bands show co-resident projection, gradient, and rasterization/supervision/loss state; the red marker gives the 9.55 GiB peak.}
\label{fig:bounded-original}
\end{figure}

Image gradients pass back through rasterization to projected attributes and then through projection to Gaussian parameters before an update.
Rasterization and loss therefore consume projected attributes and return their gradients between projection forward and backward.
The standard whole-view schedule retains regional backward states together; parameter gradients remain until the update.
Figure~\ref{fig:bounded-original} traces these states in one Original view update.

Three storage classes matter.
Gaussian parameters, Adam state, and density statistics persist across views.
Projection caches and parameter gradients are temporary view-level state, mainly scaling with participating Gaussians and parameter dimension; the full path typically allocates them for global count $N$.
Rasterization, supervision, and loss state scale with image load.
For view $v$, pixel count $P_v=H_vW_v$ drives compositing and supervision buffers, while Gaussian--tile intersection count $R_v$ drives intersection and sorting buffers.
Increasing resolution at fixed scene and field of view typically grows both.

Let $M_{\mathrm{res}}(t)$, $M_{\mathrm{view}}(t)$, and $M_{\mathrm{rast}}(t)$ denote these classes at time $t$.
Ignoring small allocations, view peak memory is
\begin{equation}
M_v^{\mathrm{peak}}
\approx\max_t\bigl[M_{\mathrm{res}}(t)+M_{\mathrm{view}}(t)+M_{\mathrm{rast}}(t)\bigr].
\end{equation}
The peak depends on concurrent state.
Spatial partitioning reduces the model floor by restricting resident Gaussians; BoundedGS retains one global model and changes the range and lifetime of the other two classes.

Given projected attributes and local pixel gradients, each region uses its own ordered Gaussian lists and pixel state for rasterization backward.
When parameters remain fixed and each region's loss can be computed locally, it can complete forward and backward computation, release local state, and contribute to one update after the complete view.
Detailed projection and gradients can likewise use Gaussians that may contribute to that view.

\subsection{Bounded execution of rasterization}

BoundedGS separates rasterization and projection backward passes, collecting projected-attribute gradients and releasing intersection, pixel, and loss caches before projection backward.
Neighboring tiles form \emph{raster leaves}, each processed sequentially through rasterization, loss, and backward computation.

Nonoverlapping leaves cover every pixel using the original camera and global coordinates.
Each leaf generates and sorts its intersections, composites pixels, computes local loss and backward contributions, and adds them to a shared projected-gradient buffer.
Its intersection, sorting, pixel, checkpoint, output, and loss state is then released.
Parameters stay fixed until one optimizer update after all leaves.

For leaf $\ell$ with $P_\ell$ pixels, weight $w_\ell=P_\ell/P_v$ preserves the full-image pixel mean of the additive $L_1$ term.
SSIM uses independent leaf windows, so its boundary treatment differs from full-image SSIM and changes with leaf layout.
Final quality is evaluated on complete images.

For leaves $\mathcal R_v$ of view $v$, let $B_\ell$ be the peak rasterization and loss state local to leaf $\ell$.
Excluding shared state and prefetch, sequential execution yields
\begin{equation}
B_v^{\mathrm{leaf}}=\max_{\ell\in\mathcal R_v}B_\ell.
\end{equation}
The largest leaf sets this part of the peak.
Uneven Gaussian coverage makes equal-area leaves unequal in load, so BoundedGS subdivides by measured demand.
Before execution, it counts Gaussian--tile intersections and combines them with pixels and buffer layout to estimate rasterization allocations without materializing or sorting the list.
Regions exceeding budget $L_{\mathrm{rast}}$ are split along tile boundaries, down to one tile.

The budget covers intersection, sorting, pixel, checkpoint, and output buffers; projection, gradients, supervision, and loss follow separate rules.
High-load areas receive smaller leaves; low-load areas retain larger ones, with one $16\times16$-pixel tile as minimum.

BoundedGS caches a CPU tree of leaf boundaries per camera.
Later visits recount current load, further split leaves above budget, and merge siblings when their combined and parent estimates pass a conservative test that is rechecked before use.
Only layout is cached, allowing adaptation as Gaussians change; resolution or budget changes rebuild the tree.
Final counts feed intersection expansion.

At higher resolution, a fixed budget distributes added pixels and intersections among more sequential leaves.
Leaves passing the budget test bound principal raster allocations, including output; local supervision, loss, and bounded prefetch remain tied to active regions.
Appendix~\ref{app:bounded-leaf} gives the byte estimate and partition rules.

\subsection{Candidate-driven projection and gradient storage}

After local rasterization reduces leaf-state lifetimes, full-model attributes, projection, and gradients can set the next peak.
Their main storage follows Gaussian count and spherical-harmonic order rather than image pixels, so smaller leaves do not reduce it.
BoundedGS uses view candidates throughout detailed projection, backward accumulation, and gradient storage.

Before detailed projection, position, opacity, maximum scale, and camera parameters yield conservative screen coverage.
Maximum scale bounds covariance independently of rotation, avoiding full rotated covariances and spherical-harmonic colors during screening.
Candidates include every Gaussian that may affect the current view under the rasterizer, retain uncertain cases, and permit false positives; detailed projection determines exact intersections.
Appendix~\ref{app:bounded-coverage} gives the construction.

For view $v$, let $\mathcal C_v$ be the candidate IDs and $n_v=|\mathcal C_v|$.
Projection reads original parameter rows by index and internally activates parameters, computes geometry, and evaluates spherical harmonics.
One global model remains; detailed forward and backward states scale with $n_v$, without a full activated tensor or parameter copy.
Screening still scans the model but avoids detailed state for every Gaussian.

Leaves share candidate projected attributes and sum rasterization-backward contributions in one projected-gradient buffer before projection backward.
Let $\boldsymbol\theta_i$ be Gaussian $i$'s parameters, $\mathbf q_i=f_v(\boldsymbol\theta_i)$ its projected attributes, and $\mathcal L_v^{\mathrm{img}}=\sum_{\ell\in\mathcal R_v}w_\ell\mathcal L_\ell$ the regional image objective.
Its parameter gradient is
\begin{equation}
\mathbf g_i^{\mathrm{img}}
=\left(\frac{\partial\mathbf q_i}{\partial\boldsymbol\theta_i}\right)^{\!\top}
\sum_{\ell\in\mathcal R_v}
w_\ell\frac{\partial\mathcal L_\ell}{\partial\mathbf q_i},
\qquad i\in\mathcal C_v.
\end{equation}
Here $f_v$ includes attribute evaluation and projection; unaffected regions contribute zero.
Accumulation does not require simultaneous leaf rasterization state.

Projection backward retains 3DGS's direct gradient accumulation (DirectGrad), writing to compact rows indexed by global Gaussian ID.
Activation derivatives are computed on that path without a separate full-model activation graph.

Final gradients for position, spherical harmonics, scale, and rotation retain candidate rows linked to global parameters by an ID map.
Parameters and Adam state keep full layouts; outside-candidate image gradients are zero.
Global opacity regularization requires a dense opacity gradient, added to image contributions.
Candidate indexing shrinks detailed projection and the retained gradient buffer, while the index and map occupy their own storage.

The optimizer reads compact gradients and IDs directly.
For rows outside the candidate set, Adam still updates the moments and parameters using a zero current image gradient.
Gradients and maps are released after the update.
Visibility is combined by union and screen gradients by accumulation before per-Gaussian densification statistics and scheduled view-level structure updates.

Tile-aligned projection groups can use the same organization.
One screening pass builds their candidate indices; each group roots a leaf tree and releases projection state after its leaf passes.
Compact gradients use the union of group candidates and accumulate repeated IDs before one update.
Grouping shrinks current projection state while earlier parameter gradients remain live; the main configuration has one full-image group.

\subsection{Resource peaks in full training}

With rendering caches controlled, structure updates and image supply can set new peaks.

Density control changes Gaussian parameters, optimizer state, and statistics together.
Rebuilding after each clone, split, parent removal, and pruning action creates intermediate layouts with concurrent old and new state.
BoundedGS instead rebuilds once after view caches are released: density control selects retained and new Gaussians, and one final mapping constructs their parameters, optimizer state, and statistics.
The CPU-assisted path holds the target layout on the host, releases the old device layout, then uploads the target, trading host memory and transfers for reduced GPU overlap.
Given the same retention and creation decisions, this changes the construction and transfer order of the layout.

For $V$ RGB images with $P_v$ pixels, retaining every decoded \texttt{uint8} image costs about $3\sum_vP_v$ bytes.
BoundedGS keeps compressed images on the host and decodes them through a fixed-capacity prefetch queue, overlapping CPU work with GPU training and bounding decoded-image residency.

Each leaf's region is packed in pinned memory, transferred as \texttt{uint8} on a separate stream, converted to floating-point supervision when used, and released after loss backward.
Prefetch holds at most the current and next leaf, tying GPU supervision memory to active regions.
Appendix~\ref{app:bounded-image-supply} details image-queue capacity and transfers.

\section{Experiments}

\subsection{Experimental Setup}
\label{sec:experimental-setup}

\paragraph{Datasets and metrics.}
The full-training comparison uses five outdoor and four indoor scenes from Mip-NeRF 360~\citep{barron2022mipnerf360}, and Barn, Family, Train, and Truck from Tanks and Temples~\citep{knapitsch2017tanks}.
Mip-NeRF 360 uses the highest-resolution images without downsampling; the Tanks and Temples training images are extracted from the original 4K videos.
The three groups have mean image dimensions of approximately $5041\times3317$, $3116\times2077$, and $3840\times2160$, denoted 5K, 3K, and 4K below.
We report PSNR, SSIM~\citep{wang2004ssim}, LPIPS~\citep{zhang2018lpips}, pure training time, and the peak temporary rendering working set.
Table~\ref{tab:main-results} averages scenes equally within each group.

\paragraph{Comparison configuration.}
The main comparison runs full-frame Original and BoundedGS in the same training framework.
Both use the absolute-gradient densification of AbsGS~\citep{ye2024absgs}, long-axis splitting from ImprovedGS~\citep{deng2025improved}, and the learnable pruning of GNS~\citep{deng2025gns}.
Densification ends at 15K steps, followed by gradual pruning to a 50\% retention target, placing the expected rendering-working-set peak near the end of densification.
The two paths share this model-count schedule, CPU-assisted structure rebuilding, and bounded JPEG image supply; each trains from SfM initialization for 30K steps once per scene.
Component and resolution experiments use short simulations at fixed Gaussian counts, while the resource ablation covers full 30K-step training.
All experiments use an NVIDIA GeForce RTX 4090.

\paragraph{Memory and timing.}
The temporary rendering working set includes projection and rasterization intermediates, supervision and loss caches, and parameter gradients.
It excludes parameters, initialized Adam state, and persistent density-control statistics.
The main table derives each scene's value at its recorded optimization-stage peak; Appendix~\ref{app:bounded-main-accounting} gives the exact accounting.
These simulations report cumulative time over all training views in their second pass.
Pure training time for the main comparison and resource ablation excludes evaluation and saving, and includes structure updates and image supply.
Structure-update memory and host memory are reported separately.

\subsection{Full-Training Quality and Cost}

\begin{table}[!htbp]
\centering
\setlength{\abovecaptionskip}{\baselineskip}
\setlength{\belowcaptionskip}{\baselineskip}
\caption{Full-training quality, time, and rendering working set. Groups equally average five Mip-NeRF 360 outdoor, four indoor, and four Tanks and Temples scenes. Time is in minutes; Mem is the scene mean of temporary working sets at recorded rendering-stage peaks, in GiB (Appendix~\ref{app:bounded-main-accounting}). Bold marks lower Mem.}
\label{tab:main-results}
\begingroup
\setlength{\tabcolsep}{6pt}
\setlength{\arrayrulewidth}{0.25pt}
\renewcommand{\arraystretch}{1.20}
\begin{tabular*}{\linewidth}{@{\extracolsep{\fill}}llrrrrr@{}}
\toprule
Dataset & Method & PSNR$\!\uparrow$ & SSIM$\!\uparrow$ & LPIPS$\!\downarrow$ & Time$\!\downarrow$ & Mem$\!\downarrow$ \\
\midrule
Mip-NeRF 360 Outdoors (5K) & Original & 24.44 & 0.739 & 0.314 & 36.7 & 9.97 \\
\rowcolor{BoundedRow}
 & \textcolor{BoundedAccent}{\textbf{BoundedGS}} & 24.42 & 0.739 & 0.314 & 33.9 & \textbf{1.76} \\
\midrule
Mip-NeRF 360 Indoors (3K) & Original & 31.52 & 0.932 & 0.239 & 16.5 & 4.43 \\
\rowcolor{BoundedRow}
 & \textcolor{BoundedAccent}{\textbf{BoundedGS}} & 31.64 & 0.932 & 0.239 & 15.9 & \textbf{0.87} \\
\midrule
Tanks and Temples (4K) & Original & 22.92 & 0.764 & 0.294 & 23.2 & 7.65 \\
\rowcolor{BoundedRow}
 & \textcolor{BoundedAccent}{\textbf{BoundedGS}} & 22.91 & 0.764 & 0.294 & 23.0 & \textbf{1.79} \\
\bottomrule
\end{tabular*}
\endgroup
\end{table}

Table~\ref{tab:main-results} shows a substantially smaller rendering working set in all three resolution groups while preserving similar final reconstruction quality.
Across the five native 5K outdoor scenes, the mean falls from 9.97 to 1.76 GiB, a reduction of 82.3\%, and training time decreases from 36.7 to 33.9 minutes.
The four 3K indoor scenes fall from 4.43 to 0.87 GiB, and the four 4K Tanks and Temples scenes from 7.65 to 1.79 GiB; their mean times decrease from 16.5 to 15.9 and from 23.2 to 23.0 minutes, respectively.
Mean PSNR changes by $-0.02$, $+0.12$, and $-0.01$ dB across the three groups, while SSIM and LPIPS remain close at the displayed precision.
Appendix~\ref{app:bounded-scene-results} gives the per-scene results.

\begin{figure}[H]
\centering
\includegraphics[width=\linewidth]{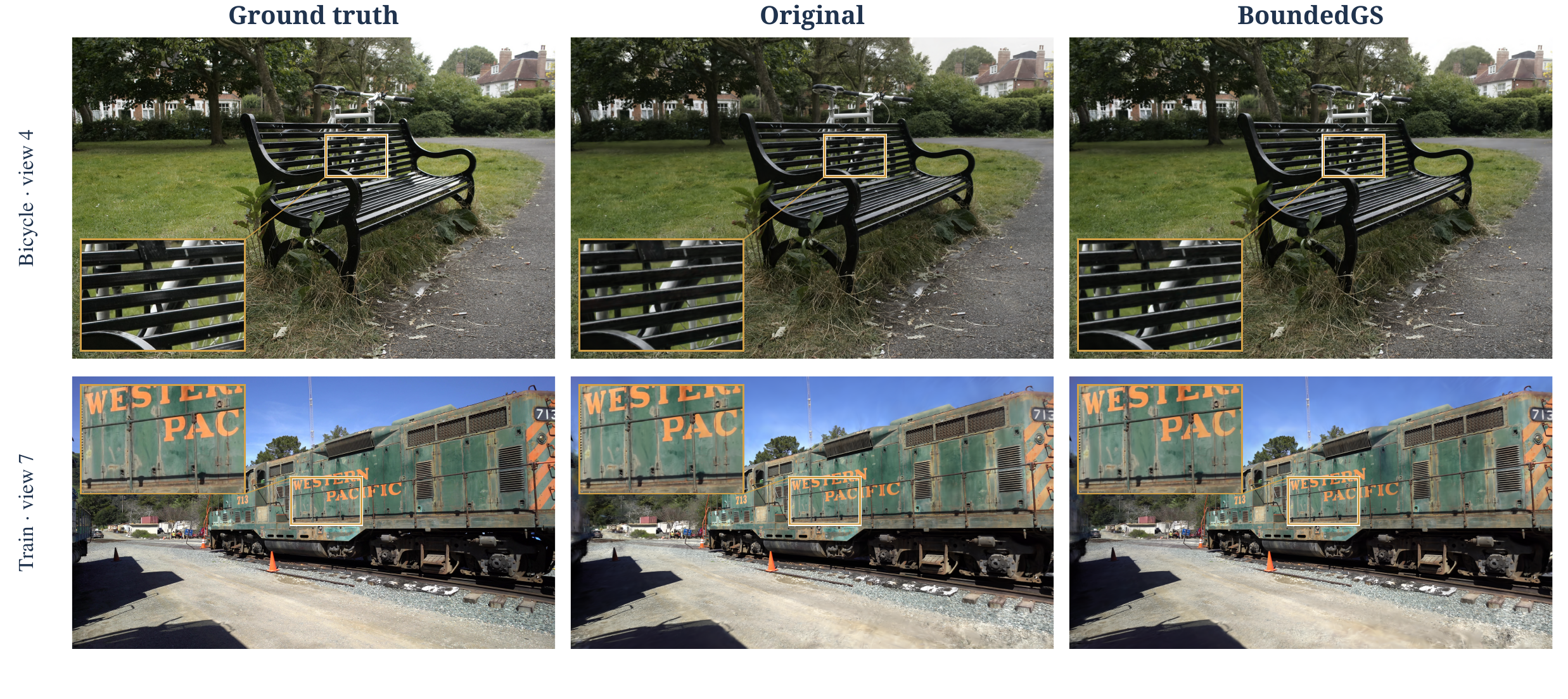}
\caption{High-resolution reconstruction. Columns show ground truth, Original, and BoundedGS; rows show Bicycle view 4 and Train view 7. Insets enlarge marked regions. BoundedGS retains similar local detail.}
\label{fig:bounded-qualitative}
\end{figure}

Figure~\ref{fig:bounded-qualitative} compares Bicycle and Train visually.
The bench slats and occlusion edges, and the lettering and panel texture on the train, are reconstructed nearly identically by BoundedGS and Original in the displayed regions.

\subsection{Rendering Components and Memory Lifetimes}

To identify the states setting each peak, we start each component variant from the same 15K SH3 model for each of five 5K outdoor scenes and add execution changes in a two-pass simulation with a fixed Gaussian count.
Table~\ref{tab:bounded-components} reports second-pass working sets and cumulative times.
$\mathrm{SSIM}_{\mathrm{Orig}}$ compares the final simulated models' renderings with Original on the same views.

\begin{table}[!htbp]
\centering
\setlength{\tabcolsep}{5pt}
\renewcommand{\arraystretch}{1.15}
\caption{Component ablation on five 5K outdoor scenes. Working set and time are second-pass scene means; $\mathrm{SSIM}_{\mathrm{Orig}}$ compares final renderings with Original, averaged over views then scenes. Rows add components sequentially; the last changes both Group layout and Leaf budget.}
\label{tab:bounded-components}
\begin{tabular}{lrrr}
\toprule
Configuration & $\mathrm{SSIM}_{\mathrm{Orig}}\uparrow$ & Working set (GiB)$\downarrow$ & Time (s)$\downarrow$ \\
\midrule
Original & 1.000000 & 9.98 & 12.19 \\
\midrule
Stage separation and early release & 0.999991 & 7.64 & 11.35 \\
\quad + Auto Leaf (384 MiB) & 0.999991 & 5.05 & 12.04 \\
\quad + Candidate-indexed projection & 0.999990 & 3.27 & 11.23 \\
\rowcolor{BoundedRow}
\quad + Compact gradients (BoundedGS384) & 0.999991 & 1.66 & 10.46 \\
\midrule
BoundedGS ($3\times3$ groups, 128 MiB) & 0.999991 & 1.31 & 12.69 \\
\bottomrule
\end{tabular}
\end{table}

\begin{figure}[!t]
\centering
\includegraphics[width=\linewidth]{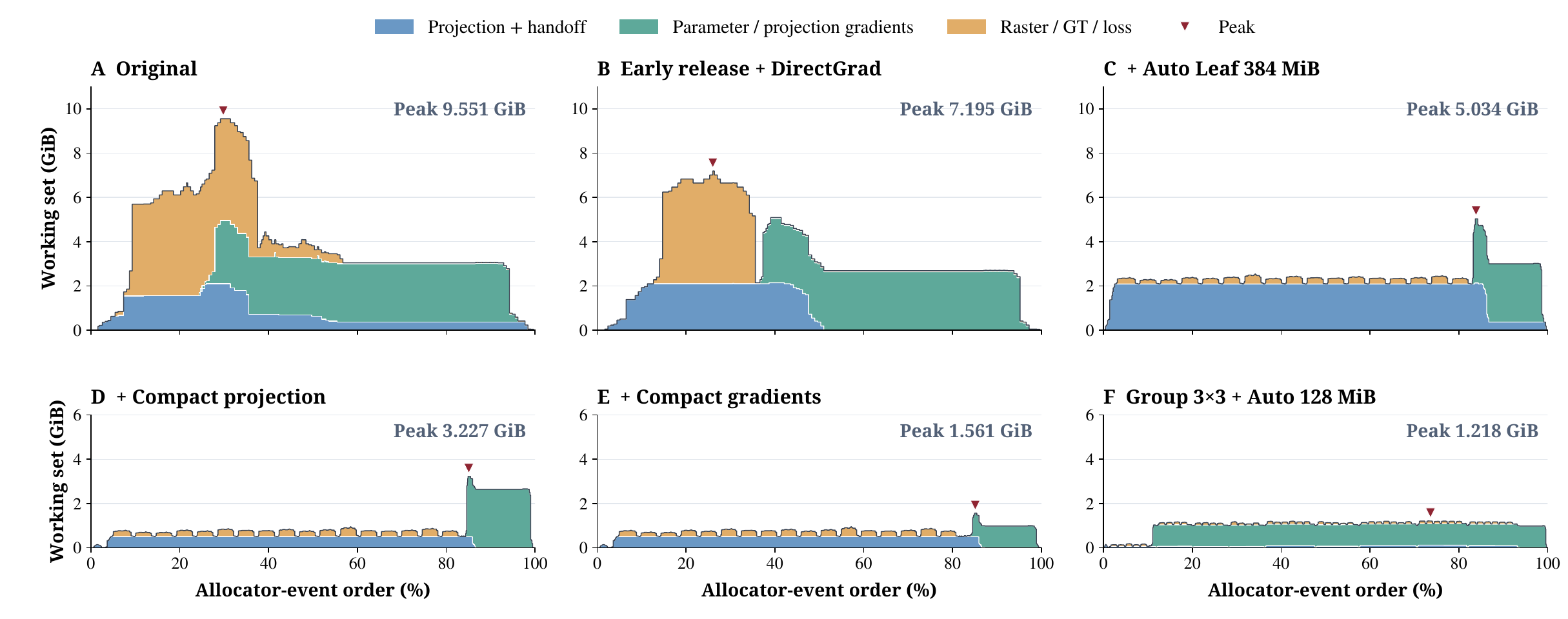}
\caption{Component working-set traces for one view of Bicycle's 15K model. A is Original; B separates backward stages; C--E add Auto Leaf, candidate projection, and compact gradients; F changes Group and Leaf budgets. All variants retain 3DGS's direct gradient accumulation. The event-order axis and stacked bands show concurrent state; red markers give peaks. A--C and D--F use different vertical scales.}
\label{fig:bounded-composition}
\end{figure}

\paragraph{Separating stages and bounding rasterization.}
Panels A--C of Figure~\ref{fig:bounded-composition} trace one Bicycle view; Table~\ref{tab:bounded-components} averages five scenes.
In Original, rasterization, projection, and parameter gradients coexist near a 9.98 GiB mean working-set peak.
B separates rasterization and projection backward, freeing raster caches before projection backward while retaining 3DGS's direct gradient accumulation.
Its raster and later gradient peaks separate, the mean working set falls to 7.64 GiB, and warm-pass time falls from 12.19 to 11.35 seconds.
C adds a load-subdivided 384 MiB Auto Leaf, reducing the mean working set to 5.05 GiB and exposing projection backward and parameter gradients as the next peak.

\paragraph{The peak moves to projection and gradients.}
Candidate-indexed projection lowers D's working set to 3.27 GiB and makes dense parameter gradients the main occupant.
E stores candidate gradients compactly, reaching 1.66 GiB and 10.46 seconds.
Relative to Original, this complete 384 MiB configuration cuts the working set by 83.3\% and time by 14.2\%; all modified variants' terminal renderings have SSIM near one relative to Original.
Rasterization, projection, and gradients successively set the peak, motivating coordinated management of their buffer lifetimes.

With $3\times3$ projection groups and a 128 MiB Leaf budget, F reduces the working set to 1.31 GiB but raises warm-pass time to 12.69 seconds.
Group-wise projection release leaves parameter gradients accumulated until the view update, forming the base in panel F.

\subsection{Resolution Scaling}

\begin{figure}[!t]
\centering
\includegraphics[width=\linewidth]{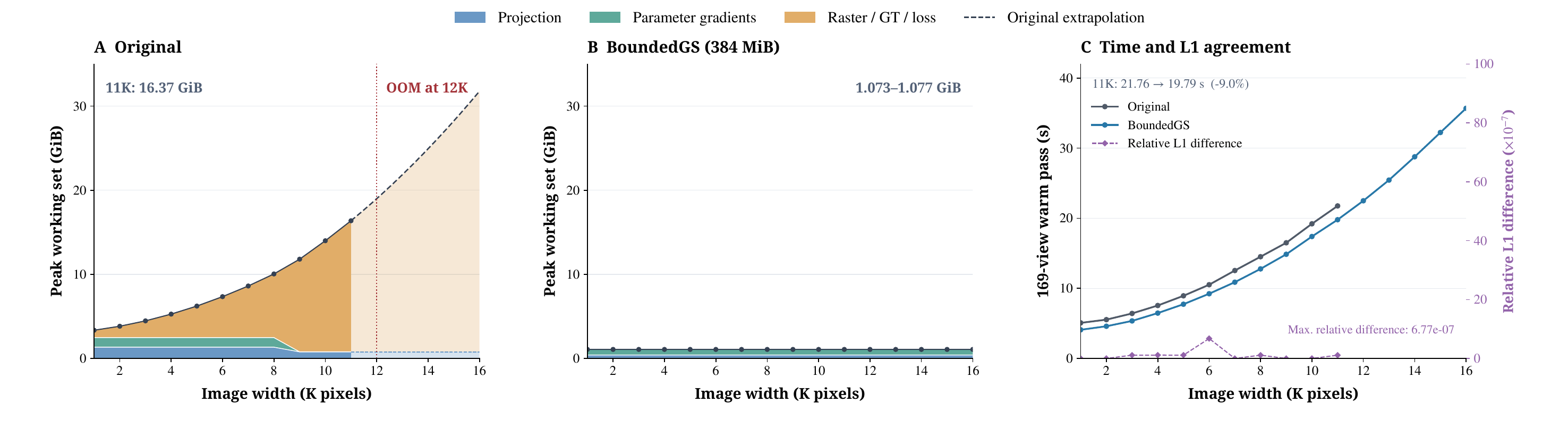}
\caption{Resolution scaling on Bicycle with the Gaussian count fixed at about 6M (1K = 1024 pixels wide). Left and center show one view's peak cache composition; right shows 169-view warm time and relative $L_1$ difference. Original is measured through 11K and reaches OOM at 12K; its higher-resolution dashed curve is extrapolated (Appendix~\ref{app:bounded-protocol}). BoundedGS is measured through 16K. Memory excludes persistent state.}
\label{fig:bounded-resolution}
\end{figure}

Figure~\ref{fig:bounded-resolution} tests resolution scaling from a common Bicycle 30K SH3 model with about 6M Gaussians, widths from 1K to 16K, constant aspect ratio and field of view, and two passes without structural updates.
Higher resolution adds pixels and Gaussian--tile intersections while projection and parameter-gradient storage remain tied mainly to the fixed Gaussian set.

Original's representative-view working set grows from about 3.36 GiB to 16.37 GiB at 11K and reaches OOM at the second 12K step.
With one projection group and a 384 MiB Auto Leaf, BoundedGS stays within 1.073--1.077 GiB over measured 1K--16K settings; maxima across all warm-pass views are 1.080--1.084 GiB.
Original's raster cache expands with image load, while BoundedGS's sequential leaves keep the working set nearly constant; Appendix~\ref{app:bounded-blocks} compares fixed-pixel blocks.
Both methods take more time at higher resolution; at the shared 11K setting, the 169-view warm pass falls from 21.76 to 19.79 seconds (about 9.0\%).
The largest relative $L_1$ difference over shared settings is about $6.8\times10^{-7}$.

\subsection{Resource Peaks in Full Training}

\begin{table}[!htbp]
\centering
\setlength{\tabcolsep}{5pt}
\renewcommand{\arraystretch}{1.15}
\caption{Full-training resource ablation on five 5K outdoor scenes. Render subtracts persistent state; Rebuild is the total allocated structure-update peak; Host is peak process RSS. All runs use the same rendering path. Memory is in GiB and time in minutes.}
\label{tab:bounded-resources}
\begin{tabular}{lrrrr}
\toprule
Configuration & Render & Rebuild & Host RSS & Time \\
\midrule
\rowcolor{BoundedRow}
BoundedGS (Full) & 1.76 & 8.68 & 13.37 & 33.95 \\
Repeated GPU rebuild & 1.76 & 12.83 & 6.30 & 30.39 \\
Predecoded host images & 1.76 & 8.68 & 17.60 & 32.19 \\
\bottomrule
\end{tabular}
\end{table}

To examine peaks outside rendering, Table~\ref{tab:bounded-resources} compares 30K Full with repeated GPU rebuilding and host-resident predecoded images on five 5K scenes; all use the same rendering path and working sets near 1.76 GiB.
Against repeated GPU rebuilding, CPU assistance lowers the structure-update peak from 12.83 to 8.68 GiB (32.4\%), raises RSS from 6.30 to 13.37 GiB, and adds 11.7\% training time.
Without it, structure updates set the GPU training peak in all five scenes, showing the need to control their layouts after rendering caches are released.
At the measured 12M SH3 budget, larger models increase the expected benefit of avoiding overlapping layouts.
Against host-resident predecoded images, bounded JPEG supply lowers RSS from 17.60 to 13.37 GiB (24.0\%) at a 5.5\% training-time cost; with similar view counts, 8K or 16K images increase the expected host-memory benefit.
The two measures exchange transfer or decoding for lower resource peaks.
\FloatBarrier

\section{Conclusion}

BoundedGS controls high-resolution working sets through load-bounded raster leaves, separated backward stages, and compact candidate state.
With the Gaussian count fixed at about 6M, it holds rendering memory within 1.073--1.077 GiB from 1K to 16K; on five native 5K scenes, full training reduces the mean working set by 82.3\% with quality close to Original and time falling from 36.7 to 33.9 minutes.

\subsection*{AI Use Statement}

Generative AI tools assisted with code writing, figure and table preparation, manuscript editing and English translation, and provided feedback on the method, experimental design, and interpretation of results.
The authors reviewed the code, analyses, figures, tables, and manuscript and take responsibility for the final content.

\subsection*{Ethics Statement}

This work studies 3D reconstruction on established benchmark datasets. We are not aware of ethical issues specific to the proposed method or its evaluation.

\subsection*{Reproducibility Statement}

Section~\ref{sec:experimental-setup} specifies the datasets and comparison protocol.
Appendix~\ref{app:bounded-protocol} documents the evaluation and memory accounting, and Appendix~\ref{app:bounded-method} provides the execution rules of BoundedGS.
Appendices~\ref{app:bounded-blocks} and~\ref{app:bounded-scene-results} report the fixed-block resource comparison and per-scene training results.

\bibliography{BoundedGS-main}
\bibliographystyle{iclr2027_conference}
\clearpage
\appendix
\section{Experimental Protocol and Accounting}
\label{app:bounded-protocol}

\subsection{Main-table runs and comparison configuration}

The main comparison covers the five Mip-NeRF 360 outdoor scenes (bicycle, flowers, garden, stump, treehill), its four indoor scenes (bonsai, counter, kitchen, room), and Barn, Family, Train, and Truck from Tanks and Temples.
Mip-NeRF 360 uses the original resolution without downsampling, and Tanks and Temples uses frames extracted from the original 4K videos.
After sorting images by filename, every eighth image is reserved for testing and the rest for training.
Original and BoundedGS each train from SfM initialization for 30K steps; training views are drawn at random without replacement from the camera pool.
Both use the absolute-gradient statistics of AbsGS, long-axis splitting from ImprovedGS, and GNS pruning: densification ends at 15K steps, then the Gaussian count is gradually reduced to 50\% retention.
The peak Gaussian budgets for the three scene groups above are 12M, 6M, and 8M.
Both paths use the same $L_1$/SSIM weights and global opacity regularizer, the CUDA Adam implementation from ImprovedGS~\citep{deng2025improved}, CPU-assisted structure rebuild, and bounded JPEG supply.
The public ImprovedGS codebase provides the densification and GNS components together with this optimizer; BoundedGS adds the execution and memory-management rules in Appendix~\ref{app:bounded-method} to its 3DGS-derived CUDA renderer.
Original performs full-frame projection and rasterization with dense parameter gradients; BoundedGS uses one projection group, a 384 MiB Auto Leaf, candidate projection, and compact gradients, as detailed in Appendix~\ref{app:bounded-method}.
PSNR, SSIM, and LPIPS are evaluated on all test views of the final 30K model, with scenes equally weighted within each group.

\subsection{Main-table training time and rendering working set}
\label{app:bounded-main-accounting}

Main-table time is synchronized wall-clock time in the training loop, including image supply, forward and backward computation, parameter updates, and structure updates, after subtracting periodic evaluation, model saving, and explicitly accounted monitor overhead.

For training iteration $i$, let $A_i$ be its optimization-stage CUDA allocated peak, $P_i$ the de-duplicated bytes of parameters and initialized Adam state at that point, and $N_i$ its Gaussian count.
Persistent density-control statistics contain one FP16 score and one int16 counter, totaling $4N_i$ bytes.
Let $i^*=\arg\max_i(A_i-P_i)$ be the recorded rendering-stage peak iteration.
The per-scene temporary rendering working set in the table is
\begin{equation}
\widehat W_{\mathrm{table}}
=\frac{A_{i^*}-P_{i^*}-4N_{i^*}}{2^{30}}\quad\mathrm{GiB}.
\label{eq:bounded-main-workset}
\end{equation}
Mem averages these values over scenes; parameter gradients, projection, rasterization, supervision images, and loss caches remain in the working set.
The resource ablation reports the total allocated peak during structure updates separately.

\subsection{Component diagnostics at fixed Gaussian counts}

The component experiments use the same 15K SH3 model within each of the five 5K outdoor scenes, hold Gaussian count fixed, and traverse training views in order twice.
Each view uses a pure $L_1$ loss and one parameter update.
The first pass establishes cached leaf layouts; the second gives cumulative time and peak temporary working set across all training views, followed by an equal-weight scene average.
Table~\ref{tab:bounded-components} starts from the standard full-frame path, then adds stage separation, Auto Leaf, candidate projection, and compact gradients; the final row combines $3\times3$ projection groups with a 128 MiB leaf budget.
After simulation, each final model and Original render the same views; image SSIM is averaged over views and then scenes and denoted $\mathrm{SSIM}_{\mathrm{Orig}}$.

\subsection{Resolution sweep at a fixed Gaussian count}

The resolution sweep starts from Bicycle's 30K SH3 model, holds its count of 5,990,000 Gaussians fixed, and uses 169 training views.
Nearest-neighbor resampling of the original images produces widths from 1024 to 16384 in increments of 1024, with rounded heights preserving the original aspect ratio and an unchanged field of view.
At each resolution, both methods start independently from the common model and zero Adam state, traverse the training views twice with pure $L_1$, and perform no structure updates.
Timing uses the second pass and excludes image decoding and resizing.
Original completes 1K--11K and runs out of memory at the second 12K step; BoundedGS completes 1K--16K.
The memory composition in Figure~\ref{fig:bounded-resolution} comes from a fixed representative view, whereas time and $L_1$ differences cover the full 169-view warm pass.
The dashed Original curve beyond 11K fits measurements from the same peak stage at 9K--11K.
For $k$ equal to image width divided by 1024, the rasterization cache is extrapolated as
\begin{equation}
M_{\mathrm{rast}}(k)=15.59+0.114(k^2-121)\quad\mathrm{GiB},
\end{equation}
anchored at 11K. Adding the 0.78 GiB projection state at that peak gives an extrapolated working set of approximately 31.76 GiB at 16K.

\subsection{Full-training resource ablation}

The resource ablation runs 30K-step training on the five 5K outdoor scenes.
All three configurations share a one-group, 384 MiB Auto Leaf, candidate-projection and compact-gradient path.
Full uses CPU-assisted one-pass structure rebuilding and bounded JPEG supply; the two comparisons substitute repeated GPU rebuilding or host-resident predecoded images.
The Render column of Table~\ref{tab:bounded-resources} subtracts concurrent persistent state, Rebuild is the structure-update allocated peak including persistent state, and Host is peak process RSS.
Time excludes evaluation and saving.

\section{Complete BoundedGS Execution Rules}
\label{app:bounded-method}

\subsection{State and execution order for one view}

Let $N$ be the current Gaussian count and $H\times W$ the full image size of the view.
Inputs include the global Gaussian parameters $\boldsymbol\theta$, camera, training image, projection-group grid side length $g$, and rasterization budget $L$ per leaf.
The image is divided into $g^2$ nonoverlapping groups; $\mathcal C_j$ denotes the screened candidate IDs of group $j$, which is then covered by raster leaves.
The main table uses $g=1$ and $L=384$ MiB, with one parameter update per training view.

For each view, the system first builds candidate indices for all groups and then processes groups in row-major order.
The current group reads the original parameter rows at its candidate IDs and projects them once for reuse by all of its leaves.
Each leaf recounts intersections under the current projection, completes local rasterization and loss backward, accumulates its contribution to the group's projected gradients, and releases its own state.
After all leaves in a group, projection backward directly accumulates parameter gradients and the group's projection and gradient caches are released before the next group.
Global regularization is computed once after all groups, and the optimizer updates after view gradients have been collected.
A Gaussian may be a candidate in multiple groups, whose disjoint pixel regions contribute to the same global parameter row.

\subsection{Conservative candidate coverage}
\label{app:bounded-coverage}

Screening reads Gaussian position, opacity, and maximum scale and estimates screen coverage using a rotation-independent scale bound.
Let $s_{i,\max}=|\gamma|\max_k\exp(s_{i,k})$ be Gaussian $i$'s maximum activated and scaled axis length and $A_i$ the projection Jacobian for the current camera.
Here $s_{i,k}$ is its $k$-th log-scale component, and $\gamma$ is the renderer's global scale modifier.
For finite inputs, screening excludes centers with camera-space depth $z_i\leq0.2$ or activated opacity $\alpha_i\leq1/255$.
It uses the detailed projection's focal lengths and clamps $x_i/z_i$ and $y_i/z_i$ to $\pm1.3\tan(\mathrm{FOV}_x/2)$ and $\pm1.3\tan(\mathrm{FOV}_y/2)$, respectively, when forming $A_i$.
The bound on its screen covariance is
\begin{equation}
\bar\Sigma_{i,2D}=A_i(s_{i,\max}^2I)A_i^\top+0.3I.
\end{equation}
For projected center $(c_{i,x},c_{i,y})$, the screen-space rectangle has half-widths
\begin{equation}
e_{i,x}=\sqrt{\max\!\left(0,2\log(255\alpha_i)\bar\Sigma_{i,2D,xx}\right)}+2,\qquad
e_{i,y}=\sqrt{\max\!\left(0,2\log(255\alpha_i)\bar\Sigma_{i,2D,yy}\right)}+2.
\end{equation}
The two-pixel additions provide the discretization margin.
The Gaussian enters a group's candidate set when $[c_{i,x}-e_{i,x},c_{i,x}+e_{i,x}]\times[c_{i,y}-e_{i,y},c_{i,y}+e_{i,y}]$ intersects its tile-padded bounds.
Nonfinite projection results retain the Gaussian for detailed projection.
Screening visits the global model and builds group candidate-ID arrays by counting and prefix sums; detailed projection and ellipse--tile intersection then follow the original rasterizer.

\subsection{Leaf budget and layout reuse}
\label{app:bounded-leaf}

A leaf is a tile-aligned rectangle $R$.
Let $n_R$ be its exact Gaussian--tile intersection count, $p_R$ its pixel count, and $t_R=\lceil h_R/16\rceil\lceil w_R/16\rceil$ its tile count for height $h_R$ and width $w_R$.
The count and offsets are obtained by ellipse--tile tests and a prefix sum.
With $u_R=\min(t_R,n_R)$, the upper bound on 32-entry backward buckets is $b_R=\lfloor(n_R+31u_R+31)/32\rfloor$.
The system evaluates the extended renderer's allocation-size queries:
\begin{equation}
E(R)=B_{\mathrm{bin}}(n_R)+B_{\mathrm{image}}(p_R,t_R)
    +B_{\mathrm{sample}}(b_R,d)+B_{\mathrm{out}}(p_R,d),
\end{equation}
where $d$ indicates whether inverse depth is enabled; the main table uses RGB without depth.
$B_{\mathrm{bin}}$ includes two uint32 Gaussian-ID arrays, two uint64 sort-key arrays, and CUB radix-sort workspace.
$B_{\mathrm{image}}$ includes per-pixel alpha and contributor state and per-tile ranges, maximum contributors, bucket counts, offsets, and scan workspace.
$B_{\mathrm{sample}}$ includes bucket-to-tile indices and transmittance and RGB state for 256 pixels per bucket; $B_{\mathrm{out}}$ includes $12p_R$ bytes of FP32 RGB output, plus $4p_R$ bytes if depth is enabled.
The same size-query routines used for allocation include 128-byte array alignment and tail padding.
Budget $L$ governs these allocations of the active leaf, while group projection and parameter gradients follow their own paths.

If $E(R)>L$ and the rectangle can be split, the system bisects it along the dimension with more tiles, breaking ties horizontally.
Children remain disjoint and cover their parent.
Subdivision stops when each leaf satisfies the budget test or has reached one tile.
If even one tile exceeds the budget, the implementation records the event and processes it at this minimum granularity.
Each final leaf is recounted under the current model and camera and follows the original rasterizer's within-tile sorting and compositing rules.

For each camera and group, a CPU tree caches the region layout; current leaf loads are recomputed at every visit.
Load increases trigger further subdivision on that visit.
When two children are leaves and both their combined estimate and the parent's estimate are at most $0.65L$, their parent is marked for possible merger on the next visit.
The proposed merger is rechecked before use; changes in image size, group grid, or budget rebuild the layout.

\subsection{Leaf loss, compact gradients, and Adam}

Let $\mathcal R_v$ be all leaves of view $v$ and $P_R$ the pixels in leaf $R$.
The main-table image loss of a leaf is $\mathcal L_R=0.8L_{1,R}+0.2(1-\mathrm{SSIM}_R)$, giving the view objective
\begin{equation}
\mathcal L_v=\sum_{R\in\mathcal R_v}\frac{P_R}{HW}\mathcal L_R+\mathcal L_{\mathrm{global}}.
\end{equation}
Area weights restore the full-image pixel mean for the $L_1$ term.
During the 15K--19K pruning window, the global term follows GNS~\citep{deng2025gns}: $\mathcal L_{\mathrm{global}}^t=\lambda_t|S_t|^{-1}\sum_{i\in S_t}z_i^t$, where $z_i^t$ is the pre-activation opacity parameter and $S_t$ is the current survivor set.
The coefficient starts at $\lambda_{15\mathrm K}=0.002$ and follows GNS's target-budget feedback; the opacity learning rate increases fivefold from $0.025$ to $0.125$ during this window.
This regularizer is applied once per view at every step of the window.
SSIM uses each leaf's own image region at its boundary.
Each leaf obtains an image gradient and explicit rasterization backward adds its contribution to the group's projected gradients; projection backward runs once after all leaves in that group finish.

Parameter gradients are allocated immediately before projection backward, avoiding their coexistence with the current group's earlier leaf state.
Compact ID set $U$ is the deduplicated union of screened candidates across groups.
An int32 map of length $N$ links global IDs to compact rows.
Position, spherical-harmonic coefficients, log scale, and quaternion use row-sparse gradients with shared candidate IDs; opacity retains a dense gradient because of global regularization.
Projection backward uses the direct gradient accumulation of 3DGS, applies activation derivatives within CUDA, and maps contributions to final candidate gradient rows.

Compact storage leaves the full parameter and Adam-momentum layout intact.
The CUDA Adam inherited from ImprovedGS uses $\beta_1=0.9$, $\beta_2=0.999$, and $\epsilon=10^{-15}$, with the same dense-gradient update semantics as PyTorch Adam; BoundedGS feeds it compact gradient rows through global IDs.
It updates every Gaussian row; a row outside the candidates has zero current image gradient while its first- and second-order momentum continue to update.
View visibility is a union of global IDs; screen gradients are accumulated by ID across groups before forming each Gaussian's densification statistic.
The main table retains ImprovedGS long-axis splitting and GNS pruning; BoundedGS organizes their rendering execution and intermediate state.

\subsection{Structure updates and image supply}
\label{app:bounded-image-supply}

A density-control event first determines retained and newly created Gaussians, then commits the structural change in one target layout.
The CPU-assisted path creates host-side target arrays for parameters, Adam state, and density statistics, moves retained rows in chunks, and appends new rows.
After releasing the old device layout, it uploads the target arrays, reducing coexistence of complete old and new GPU structures.

Training images remain as compressed JPEG bytes on the host and are decoded in camera order through a queue of capacity $K=16$.
Only the image region needed by the active leaf is uploaded asynchronously as \texttt{uint8}, converted to FP32 for loss computation, and released afterward.

\subsection{Storage composition and capacity estimate}
\label{app:bounded-capacity}

Let $D=11+3(d+1)^2$ be the parameter scalars per Gaussian at maximum spherical-harmonic order $d$.
FP32 parameters, two FP32 Adam-momentum tensors, per-Gaussian steps, and persistent density statistics occupy approximately
\begin{equation}
M_{\mathrm{resident}}(N,d)\approx\frac{(12D+28)N}{2^{30}}\quad\mathrm{GiB}.
\label{eq:bounded-resident}
\end{equation}
The main table uses SH3, where $D=59$, or about 0.685 GiB per million Gaussians; a 12M budget corresponds to about 8.23 GiB.
Equation~\eqref{eq:bounded-main-workset} gives the per-scene rendering working set, and Table~\ref{tab:bounded-resources} reports structure-rebuild peaks separately.

\section{Fixed-Pixel Blocks and Auto Leaf}
\label{app:bounded-blocks}

\subsection{Fixed-pixel blocks across resolution}

\begin{figure}[H]
\centering
\includegraphics[width=\linewidth]{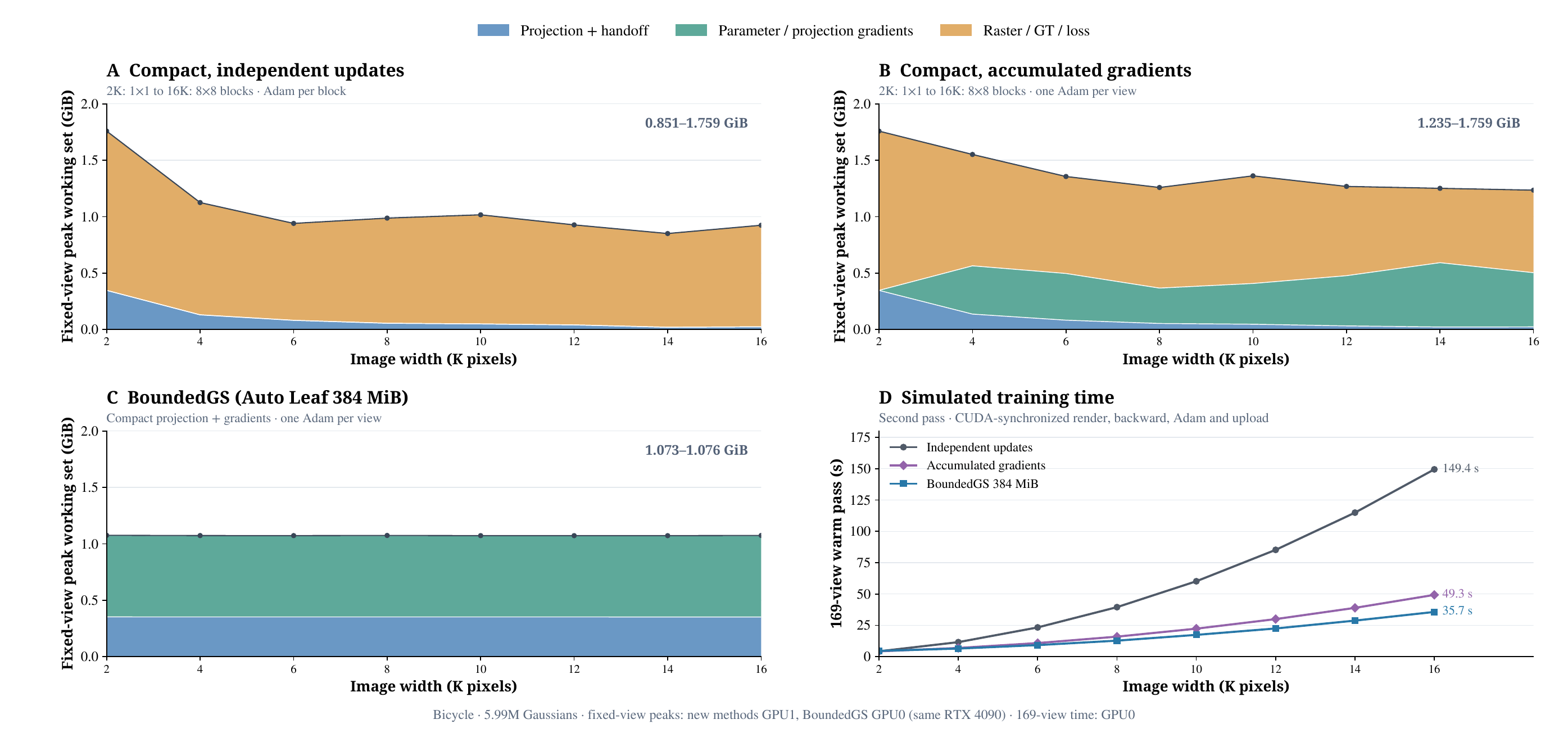}
\caption{Fixed-pixel blocks and Auto Leaf across resolution. All three paths use compact projection and gradients: A updates after each block, B accumulates gradients across fixed blocks for one update per view, and C uses a 384 MiB Auto Leaf with one update per view. The first three panels show peak-time temporary rendering state for the same representative Bicycle view; D shows synchronized second-pass time over 169 training views. Memory for A/B and C was recorded on separate RTX 4090 cards of the same model; all three times were measured on one card.}
\label{fig:bounded-pixel-block-resolution}
\end{figure}

To compare region partitioning strategies, we hold the Gaussian count of a Bicycle model fixed at about 5.99M and vary image width from 2K to 16K across three compact execution paths.
The first two use fixed-pixel blocks, either updating after every block or accumulating block gradients for one update per view.
At image width $2k\times1024$ pixels, $k=1,\ldots,8$, the fixed paths use a $k\times k$ grid: each block is 2048 pixels wide, with its height set by the fixed aspect ratio.
The third partitions by current load with a 384 MiB Auto Leaf and updates once per view.
Figure~\ref{fig:bounded-pixel-block-resolution} shows the representative-view peak composition and 169-view warm-pass time.

At 16K, independent block updates take 149.4 seconds per warm pass.
Fixed blocks with accumulated gradients use 1.24 GiB and 49.3 seconds; Auto Leaf stays within 1.073--1.076 GiB over 2K--16K and takes 35.7 seconds at 16K.
Across all measured resolutions, Auto Leaf reduces both peak rendering working set and warm-pass time relative to accumulated fixed-pixel blocks.

\subsection{Independently updated fixed grids}

\begin{figure}[H]
\centering
\includegraphics[width=\linewidth]{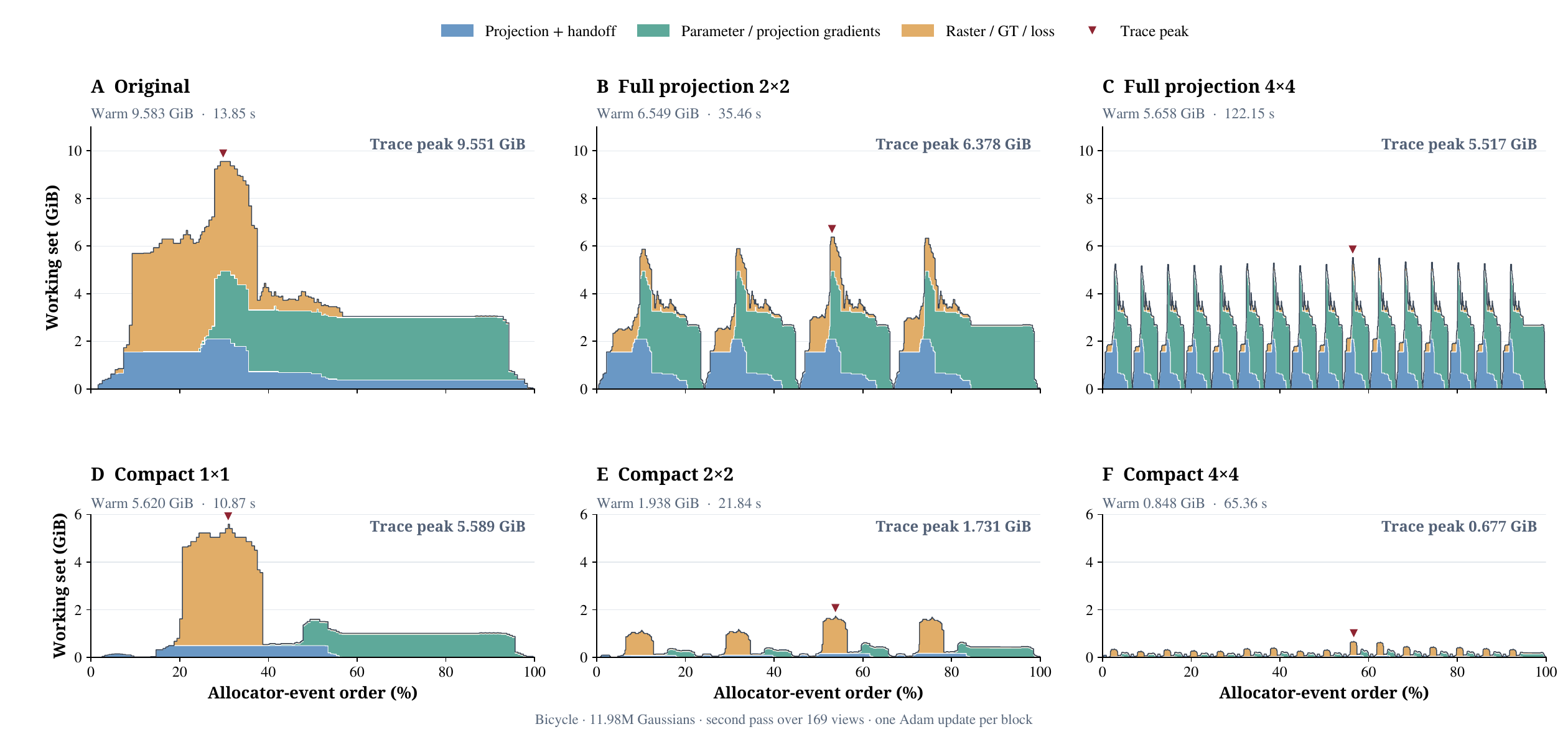}
\caption{Independently updated fixed grids on Bicycle's 11.98M-Gaussian model. The second pass spans 169 views with one Adam update per block. The upper row repeats full projection; the lower row uses stage separation, candidate projection, and compact gradients. Warm gives the all-view peak and time; Trace peak marks one view. Colors show concurrent projection, gradient, and rasterization/loss states.}
\label{fig:bounded-independent-blocks}
\end{figure}

We further use Bicycle's 15K SH3 model with about 11.98M Gaussians to examine the resource cost of independent updates on $1\times1$, $2\times2$, and $4\times4$ fixed grids.
Each block completes forward and backward computation and one Adam update.
Figure~\ref{fig:bounded-independent-blocks} reports the peak temporary working set and cumulative time over the second pass of 169 training views, together with cache traces from one representative view.

In the upper row, each image block repeats full projection.
Relative to Original's 9.58 GiB and 13.85 seconds, $2\times2$ reduces the working set to 6.55 GiB while taking 35.46 seconds, and $4\times4$ reaches 5.66 GiB while taking 122.15 seconds.
Projection and gradient state are repeatedly established across blocks in the traces.
With stage separation, candidate projection, and compact gradients in the lower row, $2\times2$ and $4\times4$ reach 1.94 and 0.85 GiB, taking 21.84 and 65.36 seconds.
On the same Bicycle model, BoundedGS Auto Leaf uses 1.57 GiB and 11.57 seconds, improving both measures over fixed $2\times2$.
With a fixed block count, pixels and intersections per block grow with resolution; subdividing within blocks according to a budget can control their rasterization caches.
Independent block updates still perform four or sixteen Adam steps per view for $2\times2$ or $4\times4$ and repeat projection and gradient construction.
Relative to BoundedGS, panel F saves a further 0.72 GiB of temporary working set atop approximately 8.21 GiB of persistent state, while requiring about 5.65 times the warm-pass time.

\clearpage
\section{Per-Scene Full-Training Results}
\label{app:bounded-scene-results}

Table~\ref{tab:bounded-scenes} reports full-training results for all 13 scenes.
BoundedGS reduces the temporary rendering working set in every scene by 72.0\%--84.9\%; PSNR differs from Original by less than 0.1 dB in 12 scenes and improves by 0.47 dB in bonsai.
Training is shorter in 11 scenes and takes 0.46 and 0.61 additional minutes in Barn and Train; mean time decreases in all three resolution groups.

\begingroup
\setlength{\tabcolsep}{4pt}
\renewcommand{\arraystretch}{1.12}
\setlength{\LTleft}{\fill}
\setlength{\LTright}{\fill}
\begin{longtable}{@{}llrrrrr@{}}
\caption{Per-scene results for Original and BoundedGS across 13 scenes. Time is pure training time in minutes and Mem is the temporary rendering working set in GiB; source values are rounded for display.}
\label{tab:bounded-scenes} \\
\toprule
Scene & Method & PSNR$\uparrow$ & SSIM$\uparrow$ & LPIPS$\downarrow$ & Time$\downarrow$ & Mem$\downarrow$ \\
\midrule
\endfirsthead
\toprule
Scene & Method & PSNR$\uparrow$ & SSIM$\uparrow$ & LPIPS$\downarrow$ & Time$\downarrow$ & Mem$\downarrow$ \\
\midrule
\endhead
\multicolumn{7}{@{}l}{\textit{Mip-NeRF 360 Outdoors (5K)}} \\*
\addlinespace[2pt]
\rowcolor{SceneTint}
bicycle & Original & 24.89 & 0.7583 & 0.2962 & 35.1 & 9.56 \\*
\rowcolor{SceneTint}
 & BoundedGS & 24.94 & 0.7585 & 0.2962 & 32.0 & 1.67 \\
\SceneSeparator
\rowcolor{white}
flowers & Original & 21.34 & 0.6205 & 0.3636 & 37.8 & 10.38 \\*
\rowcolor{white}
 & BoundedGS & 21.33 & 0.6205 & 0.3635 & 35.6 & 1.97 \\
\SceneSeparator
\rowcolor{SceneTint}
garden & Original & 26.87 & 0.8201 & 0.2119 & 40.4 & 10.20 \\*
\rowcolor{SceneTint}
 & BoundedGS & 26.87 & 0.8200 & 0.2120 & 38.1 & 2.04 \\
\SceneSeparator
\rowcolor{white}
stump & Original & 26.99 & 0.8173 & 0.3259 & 34.4 & 9.98 \\*
\rowcolor{white}
 & BoundedGS & 26.91 & 0.8165 & 0.3268 & 31.3 & 1.51 \\
\SceneSeparator
\rowcolor{SceneTint}
treehill & Original & 22.09 & 0.6811 & 0.3707 & 35.7 & 9.74 \\*
\rowcolor{SceneTint}
 & BoundedGS & 22.06 & 0.6810 & 0.3715 & 32.8 & 1.62 \\
\addlinespace[4pt]
\SceneSeparator
\multicolumn{7}{@{}l}{\textit{Mip-NeRF 360 Indoors (3K)}} \\*
\addlinespace[2pt]
\rowcolor{white}
bonsai & Original & 32.26 & 0.9436 & 0.2683 & 16.0 & 4.22 \\*
\rowcolor{white}
 & BoundedGS & 32.73 & 0.9446 & 0.2675 & 15.2 & 0.75 \\
\SceneSeparator
\rowcolor{SceneTint}
counter & Original & 29.48 & 0.9234 & 0.2472 & 15.7 & 4.03 \\*
\rowcolor{SceneTint}
 & BoundedGS & 29.48 & 0.9231 & 0.2473 & 15.4 & 0.79 \\
\SceneSeparator
\rowcolor{white}
kitchen & Original & 32.27 & 0.9317 & 0.1795 & 17.7 & 4.31 \\*
\rowcolor{white}
 & BoundedGS & 32.28 & 0.9316 & 0.1794 & 17.5 & 1.03 \\
\SceneSeparator
\rowcolor{SceneTint}
room & Original & 32.08 & 0.9282 & 0.2608 & 16.6 & 5.17 \\*
\rowcolor{SceneTint}
 & BoundedGS & 32.07 & 0.9282 & 0.2607 & 15.4 & 0.88 \\
\addlinespace[4pt]
\SceneSeparator
\multicolumn{7}{@{}l}{\textit{Tanks and Temples (4K)}} \\*
\addlinespace[2pt]
\rowcolor{white}
Barn & Original & 24.11 & 0.7386 & 0.3109 & 22.1 & 7.79 \\*
\rowcolor{white}
 & BoundedGS & 24.09 & 0.7383 & 0.3115 & 22.6 & 2.19 \\
\SceneSeparator
\rowcolor{SceneTint}
Family & Original & 23.80 & 0.7965 & 0.2569 & 24.2 & 7.60 \\*
\rowcolor{SceneTint}
 & BoundedGS & 23.85 & 0.7964 & 0.2569 & 23.5 & 1.46 \\
\SceneSeparator
\rowcolor{white}
Train & Original & 21.30 & 0.7526 & 0.2940 & 23.9 & 8.55 \\*
\rowcolor{white}
 & BoundedGS & 21.24 & 0.7522 & 0.2940 & 24.5 & 2.06 \\
\SceneSeparator
\rowcolor{SceneTint}
Truck & Original & 22.47 & 0.7702 & 0.3150 & 22.8 & 6.66 \\*
\rowcolor{SceneTint}
 & BoundedGS & 22.48 & 0.7700 & 0.3154 & 21.4 & 1.43 \\
\bottomrule
\end{longtable}
\endgroup

\end{document}